\documentclass[]{spie} 
\usepackage{cite}
\usepackage{amsmath,amssymb,amsfonts}
\usepackage{algorithmic}
\usepackage{graphicx}
\usepackage{textcomp}
\usepackage{booktabs}
\usepackage{mathtools}
\usepackage{hyperref}
\usepackage{enumerate}
\usepackage{subfig}
\usepackage{caption}
\usepackage{booktabs}
\usepackage{graphicx}
\usepackage{url}
\graphicspath{{figures/}}

\usepackage{hyperref}
\DeclareMathOperator{\inside}{\textit{inside}}
\DeclareMathOperator{\outside}{\textit{outside}}

\DeclareMathOperator{\diverge}{div}

\graphicspath{{figures/}}

\newcommand{\R}{\mathbb{R}}

\newcommand{\norm}[1]{\left\lVert#1\right\rVert}

\newcommand{\half}{\frac{1}{2}}

\begin{document}

\title{A Geometric Flow Approach for Segmentation of Images with Inhomongeneous Intensity and Missing Boundaries}

\author[a]{Paramjyoti Mohapatra}
\author[b]{Richard Lartey}
\author[a]{Weihong Guo}
\author[a]{Michael Judkovich}
\author[b]{Xiaojuan Li}
\affil[a]{Department of Mathemaitcs, Applied Mathematics and Statistics, Case Western Reserve University, Cleveland, Ohio, USA}
\affil[b]{Lerner Research Institute, Cleveland Clinic, Cleveland, Ohio, USA}

\authorinfo{Further author information: (Send correspondence to Weihong Guo)\\Weiong Guo: E-mail: wxg49@case.edu}

\pagestyle{plain}

\maketitle

\begin{abstract}
Image segmentation is a complex mathematical problem, especially for images that contain intensity inhomogeneity and tightly packed objects with missing boundaries in between. For instance, Magnetic Resonance (MR) muscle images often contain both of these issues, making muscle segmentation especially difficult. In this paper we propose a novel intensity correction and a semi-automatic active contour based segmentation approach.  The approach uses a geometric flow that incorporates a reproducing kernel Hilbert space (RKHS) edge detector and a geodesic distance penalty term from a set of markers and anti-markers.  We test the proposed scheme on MR muscle segmentation and compare with some state of the art methods. To help deal with the intensity inhomogeneity in this particular kind of image, a new approach to estimate the bias field using a fat fraction image, called Prior Bias-Corrected Fuzzy C-means (PBCFCM), is introduced. Numerical experiments show that the proposed scheme leads to significantly better results than compared ones. The average dice values of the proposed method are $92.5\%, 85.3\%,85.3\%$ for quadriceps, hamstrings and other muscle groups while other approaches are at least $10\%$ worse.
\end{abstract}

\keywords{3D segmentation, active contour, missing boundary, semi-automatic, marker, anti-marker}

\noindent \textbf{Conference topics:} Segmentation, Medical Image Analysis, Image-Based Modeling

\section{Introduction}
\label{sec:introduction}
Segmentation is an important computer vision task that has lots of practical applications. It is often used for downstream analysis.  For instance, thigh muscle morphology and composition obtained from MRI muscle segmentation have been suggested as potential imaging biomarkers for multiple diseases including osteoarthritis and sarcopenia. MRI muscle images however are very difficult to segment due to lack of boundaries between different muscle groups (physiology constraint) and intensity inhomogeneity (imaging quality constraint). Manual segmentation is time consuming and is prone to intra- and inter-operator variations. In order to receive efficient and reliable quantification of thigh muscles, robust and fast reproducible segmentation methods are desirable.  While deep learning methods have achieved considerable success when it comes to automated image segmentation, they still require a large amount of annotated data for training making them unsuitable for certain tasks. Specifically, for thigh muscles, manual annotation of data is time consuming and thus may not be feasible in all cases.

Some model based methods that incorporates auxiliary information as priors have been proposed to solve such a challenging problem.  In the past few years, several automatic methods have been proposed \cite{andrews2015generalized, baudin2012prior,chen2003using,kemnitz2017validation,lotjonen2010fast,le2016volume,mesbah2019novel,sharma2019mammogram,yokota2018automated}. Since intensity-based methods cannot distinguish different regions, most automatic methods are based on shape-based methods. Joint image processing models have attracted much attention in recent years. A segmentation model with adaptive priors from joint registration has been proposed in \cite{li2022image}. In this approach, registration plays the role of providing a shape prior to guiding the segmentation process. This method has been shown to outperform separate segmentation and registration methods as well as other joint methods. However, this method is highly dependent on the selection of moving images used for registration. When there is a significant difference between the moving image and the target image, the shape prior provided by the registration may not be accurate enough, leading to poor segmentation results. Moreover, this method requires operating on independent image pairs, and the computational cost of the energy function minimization is high. In comparison, semi-automatic methods, though require some time-consuming and tedious user intervention, are more reliable for more difficult cases. These semi-automatic methods, such as \cite{ahmad2014atlas} \cite{jolivet2014skeletal} \cite{molaie2020knowledge}, \cite{ogier2020novel}, \cite{guomarker} and \cite{ogier2017individual} have shown some efficiency in solving these problems. 

In this paper, we propose a semi-automatic method for segmenting the subgroups of thigh muscle. To the best of our knowledge, this is the first work using geometric flow with minimum user input to automatically segment 3D objects with no boundaries between objects. The novelty lies in growing an active contour propagated by an internal and external force so the segmentation results have some smoothness and stop when it is on the true boundaries. User drawn markers provide initial contours to locate objects of interest and user drawn anti-markers prevent the contours from overflowing. This is an efficient approach that allows users to feed in data adaptive information (with minimum effort) to significantly improve the segmentation results. Compared to other methods, this approach is more data adaptive and requires the least user input. One can also automate the placements of markers and anti-markers by registration but the performance is not as good as the user input ones. 

In what follows, we provide a literature review in section 2, the proposed work in section 3, and show some results in section 4. We draw conclusions and discuss future work in section 5. 

\section{Literature Review}

\subsection{Image Segmentation}

Many image segmentation techniques aim to segment the entire image domain \cite{chan2001active, mumford1989optimal}. Oftentimes, however, only a small portion of the image is of interest, and one wishes to ensure that the final segmenting contour lies within a certain region of interest.

The use of snakes to segment an image, as popularized by Kass et al. \cite{kass1988snakes}, intrinsically act as a form of selective segmentation. The idea behind snakes is to evolve a contour so that it lies on the edges of the image, thereby segmenting it. The snake, specified by the parameterization $v(s) = (x(s), y(x))$, is evolved in such a way as to minimize its energy, which, in its broadest form, is given by
\begin{equation}\label{eqn:kass}
	\begin{aligned}
		E^*_{snake} &= \int_0^1 E_{snake}(v(s)) ds
					&= \int_0^1 E_{int} (v(s)) + E_{image} (v(s)) + E_{con} (v(s)) ds.
	\end{aligned}
\end{equation}
$E_{\text{int}}$ represents the internal energy of the snake and is commonly used to penalize non-smoothness, $E_{\text{image}}$ represents energy related to image features such as edges, and $E_{\text{con}}$ encapsulates any external constraint forces that may be introduced by the user.

Caselles et al. \cite{caselles1997geodesic} expanded upon \cite{kass1988snakes} and introduced an approach that allows for the topology of the curve to change between iterations, thereby making it possible to segment multiple objects at once. This can be done by minimizing the following functional with respect to the curve $\Gamma$:
\begin{equation}\label{eqn:caselles-energy}
	E(\Gamma) = \int_\Gamma |g(\nabla I)| d\Gamma
\end{equation}
where $I$ is the image being segmented and $g(\nabla I) = \frac{1}{1+\alpha|\nabla I|^2}$ is an edge stopping function.

The final position of the snake can be further controlled by requiring the user to place a set of marker points $\mathcal{M}$ inside of a region of interest. Gout et al. \cite{gout2005segmentation} proposed a way to incorporate a set of markers into the work of Casselles et al. \cite{caselles1997geodesic} and proposed the following functional to be minimized
\begin{equation}\label{eqn:gout}
	E(\Gamma) = \int_\Gamma \mathcal{D}(x, y; \mathcal{M}) \cdot g(|\nabla I|)| ds
\end{equation}
where $\mathcal{D}(x,y; \mathcal{M})$ is the distance from a pixel $(x, y)$ to the set of markers. \eqref{eqn:gout} is close to zero if the curve $\Gamma$ either lies near the markers or on the edges of the image.

Spencer and Chen \cite{spencer2015convex} decoupled the distance penalty term from the edge stopping function and proposed minimizing 
\begin{equation}\label{eqn:spencer2015convex}
	\begin{aligned}
		F(\phi, c_1, c_2) 	&= \mu \int_{\Omega} g(|\nabla z(x,y)|)|\nabla H(\phi)| d\Omega
							&+ \lambda_1 \int_{\Omega} (z(x,y) - c_1)^2 H(\phi) d\Omega
							&+ \lambda_2 \int_{\Omega} (z(x,y) - c_2)^2 (1 - H(\phi)) d\Omega\\
							&+ \theta \int_{\Omega} \mathcal{D}(x,y; \mathcal{M}) H(\phi) d\Omega.
	\end{aligned}
\end{equation}

Typically a Euclidean distance penalty from the markers is used. Roberts et al. \cite{roberts2019convex} used a geodesic distance penalty that was based upon the edges detected in the image. This is further elaborated on in section \ref{subsec:markers_placement_geo}.

\subsection{Bias Correction}

It's not uncommon for MRI images to suffer from intensity  inhomogeneity and noise. Unfortunately, the presence of either intensity inhomogeneity and noise can make it difficult to analyze MRI images, and adversely impact the performance of many segmentation algorithms.

Assuming the intensity inhomogeneity is the result of a bias field $\beta(x)$, the observed image $f(x)$ can be modeled as follows
\begin{equation}\label{eqn:bias_model}
    f(x) = \beta(x)g(x) + \eta(x)
\end{equation}
where $g(x)$ is the true clean image and $\eta(x)$ is random Gaussian noise. By taking the logarithm of both sides of \eqref{eqn:bias_model} the model can be made to be additive.

Some of the earliest and most basic approaches to estimating $\beta(x)$ involved placing a uniform phantom inside the specific MRI images used to collect the data \cite{wicks1993correction}. This approach is undesirable, not least because it involves the need to run extra scans, but because the original MRI machine is often not available after the data has been collected. 

Another approach to correcting for the effects of the bias field involves applying a low-pass filter to $f(x)$ to estimate $\beta(x)$. The motivation behind using a low-pass filter is the assumption that the since the bias field varies slowly it will be concentrated in the lower end of the frequency domain. However, using a low-pass filter to arrive at $\beta(x)$ raises the risk that low-frequency components of the true image will get filtered out.

Pham et al. \cite{pham1999adaptive} introduced a modified form of fuzzy C-means, called adaptive fuzzy C-means (AFCM) in order to both segment the image and estimate it's bias field. The traditional fuzzy C-means objective function is 
\begin{equation}\label{eqn:fcm}
	J_{FCM} = \sum_{i,j} \sum_{k=1}^{C} u_k(i,j)^q \norm{y(i,j) - \nu_k}^2
\end{equation}
where $y$ is the observed image, $\nu_k$ is $k$-th cluster centroid, $u_k(i,j)$ is the membership value for pixel $(i,j)$, $C$ is the number of clusters, and $q$ is the weighting exponent that controls the level of fuzziness of the clustering.

\begin{equation}\label{eqn:afcm}
	\begin{aligned}
		J_{AFCM} 	= \sum_{i,j} \sum_{k=1}^{C} u_k(i,j)^2 \norm{y(i,j) - \beta(i,j)\nu_k}^2 
					+ \lambda_1 \sum_{i,j} \left( (D_i * \beta(i,j))^2 + (D_j * \beta(i,j))^2 \right)\\
	+ \lambda_2\sum_{i,j} ( (D_{ii} * \beta(i,j))^2 + 2(D_{ij} * * \beta(i,j))^2     + (D_{jj} * \beta(i,j))^2)
	\end{aligned}
\end{equation}
where $\beta(i,j)$ is the bias field at pixel $(i,j)$, $D_i$ and $D_j$ are the forward difference operators, $D_{ii}$, $D_{ij}$, and $D_{jj}$ are second-order finite difference operators, and $*$ and $**$ are, respectively, the one and two dimensional discrete convolution operators. The purpose of the second term is to minimize the variation of the bias field, and the third term penalizes discontinuities in the bias field.

Ahmed et al. \cite{ahmed2002modified} proposed a different modified version of fuzzy C-means called bias-corrected fuzzy C-means (BCFCM) that allows for the classification of each pixel to be influenced by the clustering of its neighbors. The objective function to be minimized is given by
\begin{equation}\label{eqn:BCFCM}
	\begin{aligned}
		J_{BCFCM}	&= \sum_{i=1}^C \sum_{k=1}^N u_{ik}^q \norm{y_k - \beta_k - v_i}^2
					+ \frac{\alpha}{N_R} \sum_{i=1}^{C} \sum_{k=1}^N u_{ik}^q \left( \sum_{y_r \in \mathcal{N}_k}  \norm{y_r - \beta_r - v_i}^2 \right)
	\end{aligned}
\end{equation}
where $\mathcal{N}_k$ is the set of neighboring pixels of $y_k$, $N_R$ is the cardinality of $\mathcal{N}_k$, and $\alpha$ controls the effect of the neighboring pixels on the classification of $y_k$. The added penalty term has the effect of encouraging neighboring pixels to belong to the same class, which promotes an overall classification that is piecewise-homogeneous. 

Zosso et al. \cite{zosso2017image} proposed a framework to both segment an image and estimate its bias field that is based on the Chan-Vese model for image segmentation \cite{chan2001active}. The function to be minimized is the following:
\begin{equation}\label{eqn:zosso}
	\begin{aligned}
		E_{CVB}(c_1, c_2, \phi, \beta, g)	&= \lambda_1 \int_{\Omega} (g(x) - c_1)^2 H(\phi) dx
											&+ \lambda_2 \int_{\Omega} (g(x) - c_2)^2 (1-H(\phi)) dx
											&+ \alpha \int_{\Omega} |\nabla H(\phi)| dx\\
											&+ \gamma \int_{\Omega} |\nabla \beta(x)|^2 dx
	\end{aligned}
\end{equation}
where $c_1$ and $c_2$ are constants, and the segmenting curve $\Gamma$ is given by the zero level-set of $\phi$

In addition to the above approaches, there also exists a class of methods based on Gaussian mixture models that use the E-M algorithm in order to estimate $\beta(x)$ \cite{wells1996adaptive, liu2013image}.

\subsection{ Registration Assisted  Segmentation Methods}

Given the ground truth segmentation $s$ of an atlas image $I_m$ that is not too different from the to be segmented image $I_f$, Thirion’s demons approach \cite{vercauteren2009diffeomorphic} register the two images $I_m, I_f$ and then apply the deformation field to the ground truth segmentation $s$ to achieve segmentation of $I_f$. Let $T$ be the deformation field from $I_m$ to $I_f$, one can find $T$ by solving the following minimization problem: 

\begin{equation}\label{eq:proposed}
\mathop{\mathrm{min}}\limits_T E_D(I_f, I_m\circ T)+\eta E_R(T),
\end{equation}
where $I_m\circ T$ is the deformed image of $I_m$, $E_D$ is an image dissimilarity metric that quantifies the level of alignment between $I_f$ and $I_m\circ T$, and $E_R$ is a regularization term. 

A joint segmentation-registration model was proposed by Li et al. \cite{segreg2022}. Intensity correction is also taken into consideration in the modeling. It seeks to optimize the following energy function:

\begin{equation} 
\mathcal{E}(\Theta,{u},T)=\mathcal{E}_{\mathrm{Seg}}(\Theta, { u})+\mathcal{E}_{\mathrm{CE}}(u, s \circ T)+\mathcal{E}_{\mathrm{Reg}}(\Theta, T),
\end{equation} 

\noindent where $\Theta$ collects all model related parameters, $u$ is segmentation output, $\mathcal{E}_{\mathrm{Seg}}(\Theta, { u})$ is a segmentation energy term based on  Gaussian mixture model, $\mathcal{E}_{\mathrm{Reg}}(\Theta, T)$ is a registration term moving image $I_m$ to the intensity corrected image $I_f/\beta$ with $\beta$ a bias field, and  $\mathcal{E}_{\mathrm{CE}}({ u},{ s}\circ T)$ measures the closeness between the deformed ground truth $s$ and the segmentation $u$ of the given image in terms of cross entropy.

While these methods have the obvious advantage of being fully automatic, certain disadvantages do exist. Firstly, both these methods will require an atlas segmentation map to begin with. Secondly, the atlas segmentation map must be chosen from a subject which is close to the test subject, whose segmentation is to be obtained. In other words, the deformation field between the moving and fixed image should not be large. The framework we propose in the next section will be a semi-supervised framework which will not suffer from the drawbacks that have been outlined for registration based methods. We will also compare the results obtained from our framework with the registration based frameworks.

Note that there are deep learning approaches  such as  AlexNet network \cite{ghosh2017structured}, U-Net architecture \cite{2020Clinical}, bounding boxes with 3D U-Net \cite{ni2019automatic}, and edge-aware network based on U-Net \cite{guo2021fully}. Although deep learning has great potential in muscle segmentation, we should note that this data driven method needs large amount of annotated data  and a specific network tuning for each dataset, which make it not very suitable and competitive for this task. 

\section{Proposed Work}\label{sec:proposed-work}

\subsection{Prior Bias-Corrected Fuzzy C-means (PBCFCM)}\label{sec:PBCFCM}

We propose a new method for removing intensity inhomogeneity from the thigh images that is based on a modified version of the BCFCM objective function. The new method works by incorporating prior information about the three different clusters that are present in each image into the fuzzy clustering. Let $P(\Gamma_{ki})$ denote the probability that the $k$-th pixel belongs to the $i$-th cluster. Supposing we have an estimate of $P(\Gamma_{ki})$ for all three clusters for each pixel, the proposed PBCFCM objective function is given by
\begin{equation}\label{PBCFCM}
	\begin{aligned}
		J_{PBCFCM}	&= \sum_{i=1}^C \sum_{k=1}^N u_{ik}^q \frac{1}{P(\Gamma_{ki})} \norm{y_k - \beta_k - v_i}^2
		+ \frac{\alpha}{N_R} \sum_{i=1}^{C} \sum_{k=1}^N u_{ik}^q \left( \sum_{y_r \in \mathcal{N}_k}  \norm{y_r - \beta_r - v_i}^2 \right).
	\end{aligned}
\end{equation}

Taking the partial derivatives of $J_{PBCFCM}$ and setting them equal to zero we arrive the following three updating terms:
\begin{equation}\label{eqn:u-update}
	u_{ik}^* = \frac{q}{\left(  \sum_{j=1}^c\left( \frac{D_{ik}/P(\Gamma_{ik}) + \frac{\alpha}{N_R}\gamma_i)}{D_{jk}/P(\Gamma_{jk}) + \frac{\alpha}{N_R}\gamma_j}  \right) \right)^{1/(1-q)}},
\end{equation}

\begin{equation}\label{eqn:v-update}
	v_i^* = \frac{ \sum_{k=1}^{N}u_{ik}^q \left( (y_k - \beta_k) + \frac{\alpha}{N_R} \sum_{y_r \in \mathcal{N}_k} (y_r - \beta_r) \right) }{(1+\alpha) \sum_{k=1}^{N} u_{ik}^q},
\end{equation}
and
\begin{equation}\label{eqn:beta-update}
	\beta_k^* = y_k - \frac{\sum_{i=1}^{c} u_{ik}^q v_i }{\sum_{i=1}^{c} u_{ik}^q}.
\end{equation}
where $D_{ik}$ = $\norm{y_k - \beta_k - v_i}^2$ and $\gamma_i = \sum_{y_r \in \mathcal{N}_k}  \norm{y_r - \beta_r - v_i}^2$

By thresholding the Dixon and T$_1$ high resolution images it is possible to arrive at masks for the background, fat, and muscle regions. These masks can be used to estimate $P(\Gamma_{ki})$. Suppose that it is believed that on average the three masks correctly classify a pixel with probability $\eta$, and that a misclassified pixel is equally likely to belong to one of the other two clusters. Then, the probability that a pixel is incorrectly classified by the masks is $(1-\eta)/2$. 

In general, by carefully thresholding Dixon and T$_1$ images it is possible to arrive at masks that classify most of the pixels correctly. This is because despite being a low-resolution image, the Dixon image is relatively free of intensity inhomogeneity and can be used to create an accurate fat mask. The background mask can be obtained by thresholding the T$_1$ image with a value close to zero, and the muscle mask can be obtained by selecting all of the pixels that are not in either the fat or background masks.

The steps of the PBCFCM algorithm can be summarized as follows:
\begin{enumerate}
    \item Determine the masks for the three clusters by thresholding the Dixon and T$_1$ images.
    \item Use the three masks to estimate $P(\Gamma_{ki})$.
    \item Initialize the cluster centroids and set the initial bias field to be close to be close to zero.
    \item Update $u$, $v$, and $\beta$ using \eqref{eqn:u-update}, \eqref{eqn:v-update}, \eqref{eqn:beta-update}.
    \item Repeat steps 3 and 4 until the vector of cluster centroids, $v$, converges.
\end{enumerate}

We note that this intensity correction step might not be necessary for other data or a different intensity correction algorithm might need to be adopted. 

\subsection{RKHS Edge Detection}\label{sec:RKHS}

In order to do a better job at identifying the weak edges in-between the muscle groups we use an RKHS edge detector \cite{burrows2020reproducible, deng2016single}.

Let $z$ be a function on a continuous domain $\Omega = [0,1] \times [0,1]$ that gives the intensity values for a 2D image. Then, the following Gaussian kernel can be used to model the smooth regions of the image:
\begin{equation}\label{eqn:RKHS}
	\mathbb{K}(x, \tilde{x}) = \left(\frac{1}{\sqrt{2\pi}\sigma}\right)^2 e^{-\frac{|x - \tilde{x}|^2}{2\sigma^2}}.
\end{equation}

Smooth approximations to the Heaviside function
\begin{equation}\label{eqn:app-heav}
	\psi(x) = \half \left( 1 + \frac{2}{\pi}\arctan\left(\frac{x}{\epsilon}\right)  \right).
\end{equation}
can be used to form a redundant dictionary, allowing for the edges of the image to be modeled using the following function\begin{equation}\label{eqn:model-edges}
	h(x) = \sum_{i=1}^{l}\sum_{j=1}^{N} b_{ij}\psi((\cos\theta_i, \sin\theta_i) \cdot x + c_j)
\end{equation}
where $l$ is the number of orientations for which an edge is checked, $N$ is the total number of pixels in the image, $b_{ij}$ is the edge weight, $c_j \in \{0, \frac{1}{N-1}, \frac{2}{N-1}, \ldots, 1\}$ is the position of the pixel, and $\theta_i$ is the orientation of the edge. Since $l$ is the number of orientations for which an edge is checked, we get that $\theta_i \in \{0, \frac{2\pi}{l}, 2(\frac{2\pi}{l}), \ldots, (l-1)(\frac{2\pi}{l}) \}$.

Combining \eqref{eqn:RKHS} and \eqref{eqn:model-edges}, z can be approximate by $f$ according to
\begin{equation}\label{eqn:RKSH-Heavi}
	f(x) = \sum_{i=1}^N d_i \mathbb{K}(x, x_i) + \sum_{i=1}^l \sum_{j=1}^N b_{ij} \psi((\cos\theta_i, \sin\theta_i) \cdot x + c_j).
\end{equation}

If we let $\vec{f} = (f(t_1), f(t_2), \ldots, f(t_N))'$ represent a discretization of $f$ on the grid $t_i = (x_i, y_i) \in [0,1] \times [0,1]$, where $i = 1,2,\ldots,N$. Using matrices, the image $\vec{f}$ (whose pixels are arranged lexicographically in a single column) can be written as $\vec{f} = Kd + \Psi \beta$ where $K_{mn} = \mathbb{K}(x_m, x_n)$ and $\Psi$ contains the values of $\psi((\cos\theta_i, \sin\theta_i) \cdot x + c_j)$ arranged into a matrix, and $\beta$ is a column vector containing each $b_{ij}$ in lexicographical order. Now suppose that $\vec{z}$ is a discretization of an analog image $z$ which we wish to approximate using $K_{mn}$ and approximated Heaviside functions. Then, $d$ and $\beta$ can be determined via the following minimization model:
\begin{equation}\label{eqn:RKHS-min}
\begin{aligned}
    \min_{d,\beta} \half \norm{\vec{z} - (Kd + \Psi \beta)}^2 & + \gamma d^TKd + \alpha\norm{\beta}_1
    + \nu g^T |\nabla (Kd+\Psi \beta)|,
\end{aligned}
\end{equation}
where $\gamma$, $\alpha$, and $\nu$ are constants, and $g = g(\Psi \beta)$ is an edge stopping function. The purpose of the first term in \eqref{eqn:RKHS-min} is to ensure that the approximation, $Kd + \Psi \beta$, closely matches the original discretized image $\vec{z}$. The second term penalizes non-smoothness, and the third term promotes the sparsity of $\beta$ since $\{\psi((\cos\theta_i, \sin\theta_i) \cdot x + c_j)\}$ is a redundant dictionary. Finally, the last term penalizes low contrast near the edges and high contrast in smooth regions. The minimization of \eqref{eqn:RKHS-min} can be carried out efficiently using the alternating direction method of multipliers (ADMM).

\subsection{Marker Placement and Geodesic Distance Maps}\label{subsec:markers_placement_geo}

Since the edges in-between the muscle regions are often too weak to pick up (even using the RKHS edge descriptor) we require the user to draw both markers and anti-markers for each of the three muscle groups. The purpose of the markers is to be able to identify which group is being segmented. The anti-markers are required in order to fill in for the weak edges and to prevent the segmenting contour from leaking into the other two muscle groups.

The markers are drawn in the form of a polygon using the \texttt{roipoly} command in MATLAB. Ideally, the markers should coincide with each of the muscles in the group being segmented, and should not overlap with any of the other muscles.

Just like the markers, the anti-makers are also drawn in the shape of a polygon, and they should overlap with all of the muscles that do not belong to the group being segmented. While neither the markers or the anti-markers need to be drawn with a high level of precision, the anti-markers should be drawn carefully along the boundary of the muscle group being segmented in order to help fill in for the missing edges.

In order to incorporate the markers into the geometric flow (see section \ref{subsec:geo-flow}), it is necessary to compute distance maps from the markers and anti-markers.  This can be done efficiently by using the fast sweeping method \cite{zhao2005fast} to compute the viscosity solution to the Eikonal equation:
\begin{equation}\label{eqn:eikonal}
	\begin{aligned}
		|\nabla \mathcal{D}(x)| &= f(x), \quad x \in \R^n\\
		\mathcal{D}(x) &= \phi(x) \quad x \in \mathcal{M} \subset \R^n
	\end{aligned}
\end{equation}
where $\mathcal{D}(x)$ is the distance of point $x$ from the set of markers $\mathcal{M}$. Note that if $\phi(x) \equiv 0$ and $f(x) \equiv 1$, then $\mathcal{D}$(x) is simply the Euclidean distance.

In order to arrive at the geodesic distance which incorporates information about the edges in-between the muscles, the distance from the user drawn markers $D_{\mathcal{M}}$ ($\mathcal{D}_{\mathcal{AM}}$ is defined similarly) is given by 
\begin{equation}\label{eqn:eikonal-rkhs}
	\begin{aligned}
		|\nabla \mathcal{D}_{\mathcal{M}}(x)| &= \epsilon + \omega |\Psi \beta|, \quad x \in \R^n\\
		\mathcal{D}_{\mathcal{M}}(x) &= 0, \quad x \in \mathcal{M} \subset \R^n
	\end{aligned}
\end{equation}
where $\Psi \beta$ is the same as in \ref{sec:RKHS},  $\omega > 0$ and $\epsilon$ is a small positive constant that ensures the distance between any two different points is non-zero.

Once the distance maps have been obtained for both the markers and the anti-markers, they can be combined to form a single distance penalty
\begin{equation}\label{eqn:dist_penalty}
	\mathcal{D}_P = \half \left( \mathcal{D}_{\mathcal{M}} + \frac{e^{-\alpha \mathcal{D}_{\mathcal{AM}}} - e^{-\alpha}}{1 - e^{-\alpha}} \right).
\end{equation}
that will be incorporated into the segmentation algorithm.

If a 3D version of the fast sweeping method is used to solve the Eikonal equation then it is only necessary to draw markers and anti-markers on only a fraction of the slices. The exact number will depend upon the resolution of the images in the $z$ direction and the level of noise.

\subsection{Marker Placement Using Atlas-Based Segmentation}\label{sec:markers}

It is possible to cut the amount of user input in half by using atlas-based segmentation to draw the markers. Suppose that the ground truth segmentation is available for a template image $T$, and we wish to segment a reference image $R$. Let $\mathcal{M}$ be the 3D binary mask that represents the known segmentation for $T$. Registering the template and reference images we arrive at a transformation $\phi(x)$ such that $T(\phi(x)) \approx R(\phi(x))$. Then, the markers $\hat{M}$ for the reference image $R$ can be taken as
\begin{equation}
    \hat{\mathcal{M}} = \phi(\mathcal{M}).
\end{equation}

While it is also technically possible to automate the placement of the anti-markers in the exact same way, doing so often leads to poor segmentation results. This is because the anti-markers help to fill in for the missing edges, and they need to be drawn more precisely than the markers need to be drawn. 

\subsection{Segmentation Using Geometric Flow}\label{subsec:geo-flow}

In order to represent the closed curve $C$ that defines the segmentation we adopt the level-set method that was introduced by Osher and Sethian \cite{osher1988fronts}. In other words, $C$ is defined as the zero level-set of a Lipschitz  function $\phi(x)$ 
\begin{equation}\label{eqn:level_set_method}
	\begin{cases}
		C &= \{x : \phi(x) = 0\},\\
		\inside(C) &= \{x : \phi(x) > 0\},\\
		\outside(C) &= \{x : \phi(x) < 0\}.
	\end{cases}
\end{equation}

$\phi(x)$ is initialized to be 1 inside of several user placed seeds, and -1 everywhere else. 

After being intialized, $\phi(x)$ is propagated along the normal direction with velocity $F$ according to the following partial differential equation \cite{guo2008geometric}
\begin{equation}\label{eqn:d_phi/d_t}
	\frac{\partial \phi}{\partial t} = (F + q \kappa)|\nabla \phi|
\end{equation}
where $q>0$ and $\kappa = \diverge \left(\frac{\nabla \phi}{|\nabla \phi|}\right)$ is the curvature of $\phi$. Using Euler's method, $\phi$ is updated at time $t$ as follows:
\begin{equation}\label{eqn:phi_update}
	\phi_t(x) = \phi_{t-1}(x) + h(F+q\kappa)|\nabla \phi(x)|
\end{equation}

While $F$ is typically taken to be an edge stopping function (i.e. $g = \frac{1}{1+\alpha|\nabla f(x)|}$), using such a simple force is insufficient for muscle segmentation. Rather, we define a new force
\begin{equation}\label{eqn:F}
    F = g - \gamma \mathcal{D}_P - \eta I
\end{equation}
where $g = \frac{1}{1+\alpha|\nabla \Psi \beta|}$ is an edge stopping function that uses the edge descriptor $\Psi \beta$, $\mathcal{D}_P$ is the distance penalty calculated in section \ref{sec:markers}, $\gamma, \eta >0$ are constants, and $I$ is an intensity fitting term which is defined below.

The data fitting term has a similar function as the fidelity terms in the Chan-Vese model. Suppose $\xi$ is the average intensity of the of pixels inside $C$ prior to the first iteration of the algorithm. Then, if $z$ is the bias corrected image,
\begin{equation}
	I(x) = 
	\begin{cases}
		|z(x) - \xi| & \text{ if } z(x) - \xi < 0\\
		\psi |z(x) - \xi| & \text{ if } z(x) - \xi \ge 0 \\
	\end{cases}
\end{equation}
where $0 < \psi < 1$ helps to address the asymmetry between the average differences between muscle and fat, and muscle and background intensity values.

Note that the only positive contribution to the force comes from the edge stopping function $g$. While the edge stopping function is bounded between zero and one, it isn't possible to place rigorous bounds on $\gamma \mathcal{D}_P$ and $\eta I$. Consequently, we round and value of $F$ that are less than -1 up to -1 so that $-1 \le F < 1$. Additionally, a median filter is also applied to $F$ to remove outliers.

\subsection{Summary of Proposed Approach}
The follow list of steps summarizes the proposed segmentation algorithm:
\begin{enumerate}
	\item Perform intensity correction using PBCFCM. Note this step can be skipped or changed for different data.
	\item Calculate $\Psi \beta$ according to \eqref{eqn:RKHS-min}.
	\item Draw the markers and anti-markers and calculate the distance penalty $\mathcal{D}_P$ for each muscle group. 
	\item Assemble $F$ according to \eqref{eqn:F} and perform the segmentation using geometric flow for each of the muscle groups.
\end{enumerate}

\section{Results}

We apply the proposed segmentation algorithm from section \ref{sec:proposed-work} to six different subjects and compare with some related work. In addition, we also assess the performance of the newly proposed PBCFCM approach to removing intensity inhomogeneity against that of BCFCM \cite{ahmed2002modified} and LGMM \cite{liu2013image}. The performance of the proposed segmentation algorithm is tested on all three thigh muscle groups for six different subjects. The high resolution T$_1$ images were collected using a 3T MRI scanner (SkyraFit, Siemens) with an 18-channel flex coil. The TR/TE was 795/10 (ms), the field-of-view (FOV) was $400 \times 312 \times140$ mm$^3$, and the image size was $512 \times 299 \times 28$. For the fat fraction images the TR/TSE was $16.37/0.94, 2.05, 3.16, 4.27, 5.38, 6.49$, the FOV was $400 \times 300 \times 140$ mm$^3$, and the image size was $256 \times 191 \times 28$.

\subsection{Intensity Correction}

\begin{figure} [ht]
   \begin{center}
   \begin{tabular}{c} 
   \includegraphics[height=10cm]{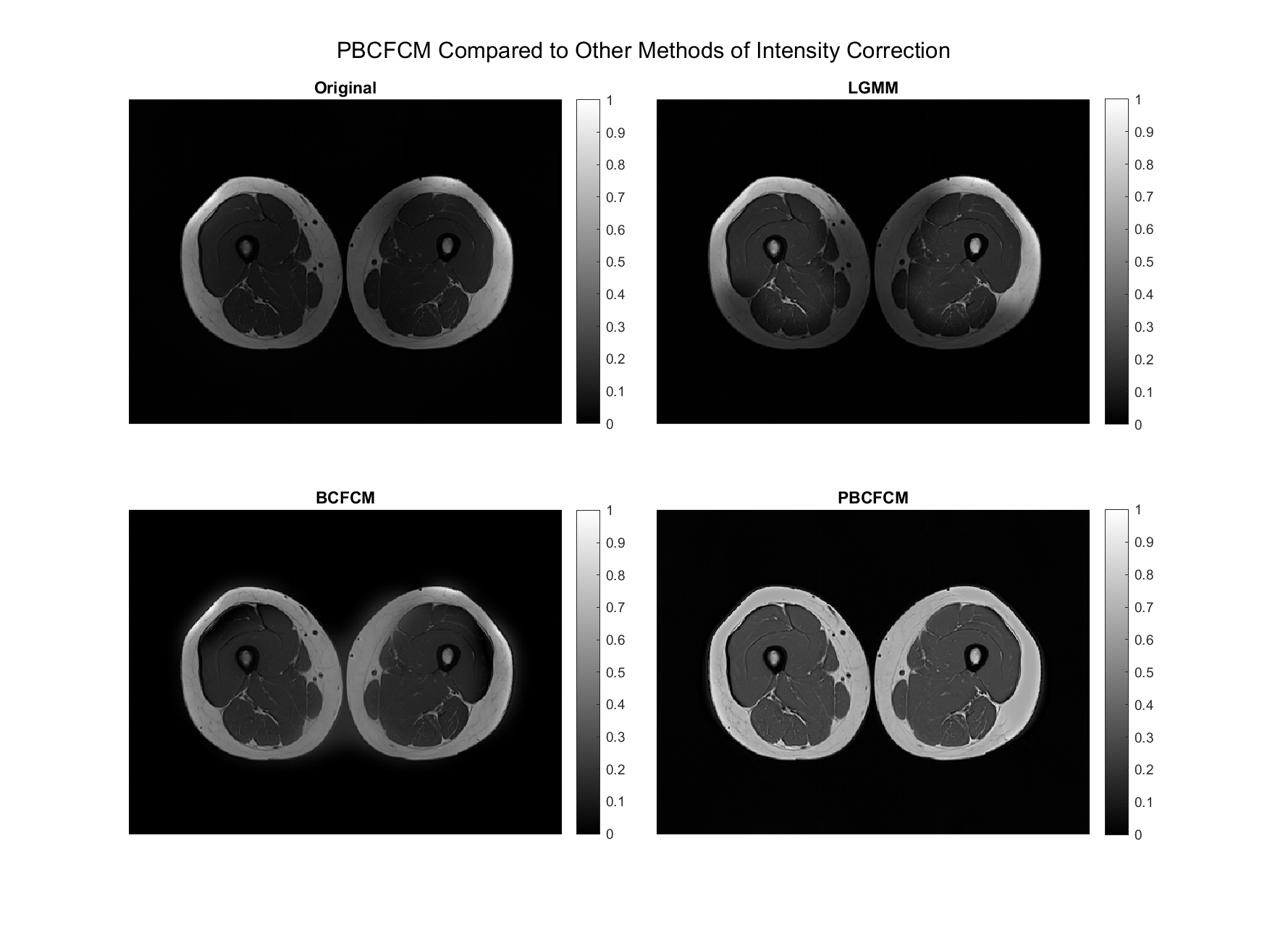}
   \end{tabular}
   \end{center}
   \caption[example] 
   { \label{fig:PBCFCM_DEMO} 
A comparison between the newly proposed method for intensity correction (bottom right), LGMM (top right), and BCFCM (bottom left).}
   \end{figure}

Unfortunately, since there is no way to obtain a ground truth MRI image with no intensity inhomogeneity there is no simple way to quantifying PBCFCM's performance. Figure \ref{fig:PBCFCM_DEMO} provides a visual comparison between the proposed method, BCFCM, and a Local Gaussian Mixture Model (LGMM) based approach from which it can be seen that PBCFCM yields the best result. The original image is shadowed, and the tops of the quadriceps appear to be illuminated. The LGMM is able to do a good job correcting the top half of the image, but the bottom half still suffers from the effects of intensity inhomogeneity. The BCFCM corrected image looks better than the LGMM image, but there are still significant levels of intensity inhomogeneity present within the muscle portions of the image. The PBCFCM corrected image, on the other hand, has much better contrast, and the image is much closer to being piece-wise constant . It's also worth noting that the image's finer details, in particular the intramuscular fat, aren't washed out by the correction process.

\begin{figure} [ht]
   \begin{center}
   \begin{tabular}{c} 
   \includegraphics[height=10cm]{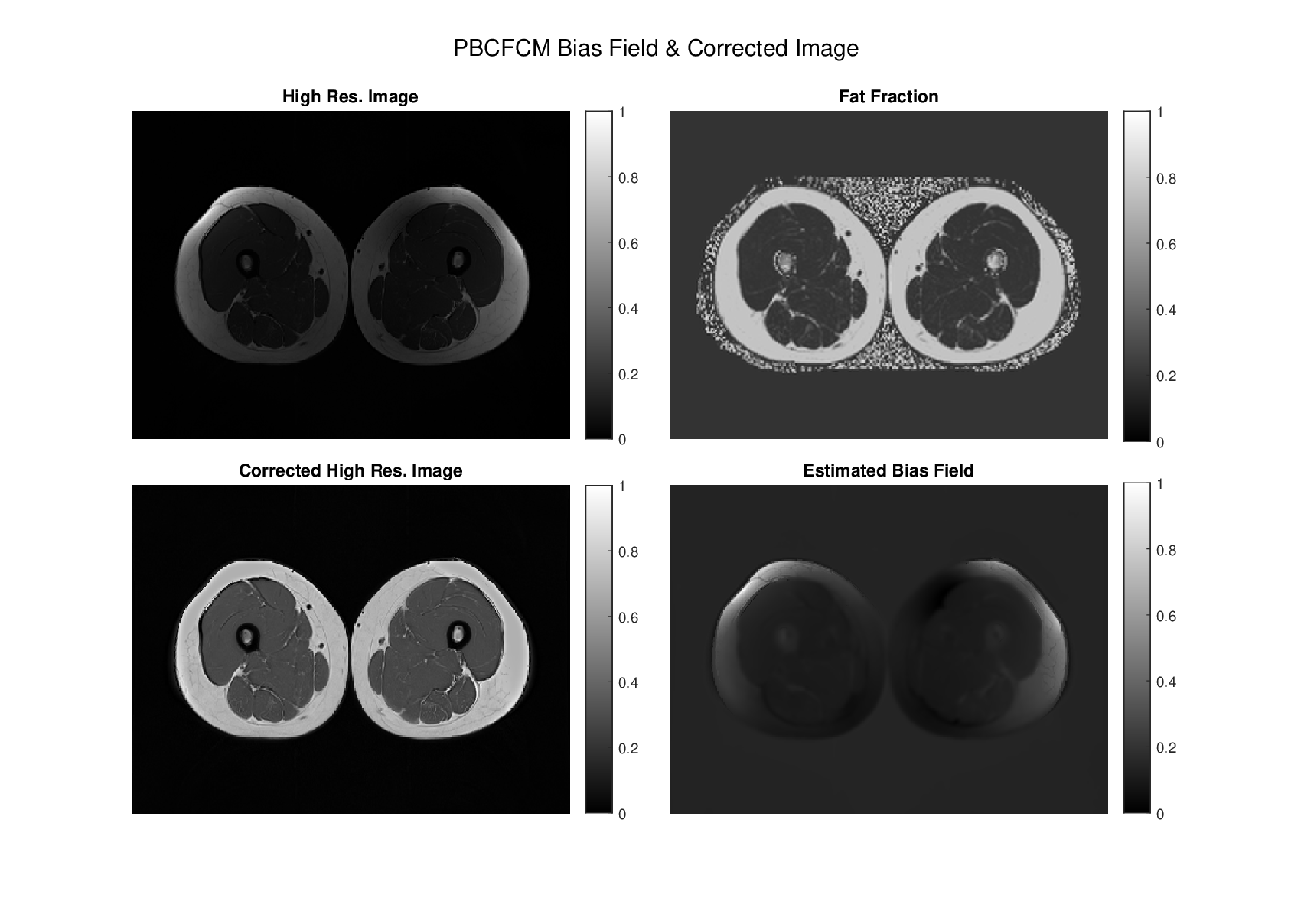}
   \end{tabular}
   \end{center}
   \caption[example] 
   { \label{fig:PBCFCM_Bias} 
The original high resolution image (top left), the fat fraction image (top right), the PBCFCM corrected high resolution image (bottom left), and the estimated bias field (bottom right).}
   \end{figure}

Figure \ref{fig:PBCFCM_Bias} provides an example of a high resolution image next to its corresponding fat fraction image. As described previously in \ref{sec:PBCFCM}, both of these images are used to create the three masks that are needed for PBCFCM. While the fat fraction image contains much less intensity inhomogeneity than the high resolution image, it isn't practical to use it to perform the segmentation because its resolution is too low. Figure \ref{fig:PBCFCM_Bias} also shows the estimated bias field, $\beta$, for the original image, which is subtracted from the original image in order to get the PBCFCM corrected image.

Another way of evaluating the performance of PBCFCM is by observing it's impact on other image processing tasks such as segmentation and registration. Section \ref{seg:segmenation_results} presents the Dice values for the proposed approach, all of which are greater than 85\%. The proposed segmentation algorithm greatly depends on the image's intensity values, and if PBCFCM isn't used to correct the intensity values prior to performing the segmentation then it is difficult to obtain meaningfully accurate results. For the subjects that exhibit the most intensity inhomogeneity the accuracy of the segmentation falls below 50\% if PBCFCM is not applied, which further demonstrates the value that PBCFCM provides.

\subsection{Segmentation Results}\label{seg:segmenation_results}

For the sake of consistency, the markers and anti-markers for all six of the subjects were drawn by the same person. Out of the 28 slices, markers and anti-markers were drawn on slices 3, 8, 13, 18, 23, and 26. For the Eikonal equation \eqref{eqn:eikonal-rkhs}, $\epsilon = 0.1$ and $\omega = 10^4$ were used. In order to define the force $F = g - \gamma D_P - \eta I$ \eqref{eqn:F} for the geometric flow, $\gamma$ and $\eta$ were both chosen to be between five and ten depending on the subject and the muscle group being segmented. Larger values of $\eta$ and $\gamma$ were chosen when the edges in-between the muscle groups were more clearly defined, while smaller values were used when the edges were more blurred.

The accuracy of the segmentation was quantified using the Dice coefficient which is given by
\begin{equation}
	DSC = \frac{2|X \cap Y|}{|X| + |Y|}
\end{equation}
where $X$ and $Y$ are two sets, and $|\cdot|$ denotes the cardinality of a set. Table \ref{tab:Dice} shows the Dice values of the proposed scheme for each of the three different muscle groups for all six subjects, along with the averages for each of the three groups across all of the subjects. 

\begin {table}
\caption {Dice Values of the proposed method with Hand-Drawn Markers } \label{tab:Dice} 
\begin{center}
	\begin{tabular} {p{.50in} p{.45in} p{.45in} p{.45in}}
		\toprule[1.5pt]
		{\bf PID} & {\bf Quads} 	& {\bf Hams} 	& {\bf Others}\\ 
		\midrule
		101   &  95.25\%    & 88.04\%   & 89.41\% \\
		102   &  92.00\%    & 88.80\%   & 90.57\% \\
		103   &  90.53\%    & 85.71\%   & 74.28\% \\
		104   &  90.34\%    & 79.36\% 	& 79.73\% \\
		105   &  93.95\%    & 85.59\%   & 86.87\% \\
            106   &  92.73\%    & 84.45\%   & 90.78\% \\
		\midrule
 		\textbf{AVG}    &\textbf{92.47\%}   &\textbf{85.33\%}		&\textbf{85.27\%}	\\
		\bottomrule[1.25pt]
		\end {tabular}
	\end{center}
\end {table}
\begin {table} [!htbp]
\caption {Dice Values - Image registration algorithm (M1)} \label{tab:M1} 
\begin{center}
	\begin{tabular} {p{.50in} p{.45in} p{.45in} p{.45in} p{.45in}}
		\toprule[1.5pt]
		\textbf{Fixed PID} & \textbf{Quads}	&  \textbf{Hams} &  \textbf{Others}		&\textbf{Moving PID}\\ 
		\midrule
		101 &  80.53\%    & 77.43\%   & 72.97\%  & 102 \\
		102 &  73.59\%    & 73.38\%   & 63.04\%  & 101 \\
		103 &  71.75\%    & 80.13\%   & 69.25\%  & 102 \\
		104 &  50.83\%    & 65.72\%   & 43.71\%  & 102 \\
		105 &  81.65\%    & 74.38\%   & 73.75\%  & 102\\
            106 &  80.19\%    & 82.34\%   & 72.89\%  & 102\\
		\midrule
		\textbf{AVG}    &\textbf{73.09\%}   &\textbf{75.56\%} &\textbf{65.93\%} & \\
		\bottomrule[1.25pt]
		\end {tabular}
	\end{center}
\end {table}
\begin {table}[!htbp]

\caption {Dice Values - Simultaneous Seg-Reg Algorithm (M2)} \label{tab:M2} 
\begin{center}
	\begin{tabular} {p{.50in} p{.45in} p{.45in} p{.45in} p{.45in}}
		\toprule[1.5pt]
		{\bf Fixed PID} & {\bf Quads} 	& {\bf Hams} 	& {\bf Others}		&\textbf{Moving PID}	\\ 
		\midrule
		101       &  82.96\%    	& 70.01\%      	& 60.57\% 			& NA				\\
		102       &  77.76\%    	& 64.30\%      	& 58.84\% 			& NA				\\
		103       &  16.47\%    	& 5.70\%      	& 5.51\% 			& NA				\\
		104       &  43.51\%    	& 60.31\%      	& 20.05\% 			& NA				\\
		105       &  23.15\%    	& 5.25\%      	& 14.3\% 			& NA				\\
  		106       &  32.86\%    	& 5.65\%      	& 16.71\% 			& NA				\\
		\midrule
		\textbf{AVG}		  &  \textbf{46.12\%}		& \textbf{35.20\%}		& \textbf{29.33\%	}	& 	\\
		\bottomrule[1.25pt]
		\end {tabular}
	\end{center}
\end {table}

   \begin{figure} [!htbp]
   \begin{center}
   \begin{tabular}{c} 
   \includegraphics[height=5cm]{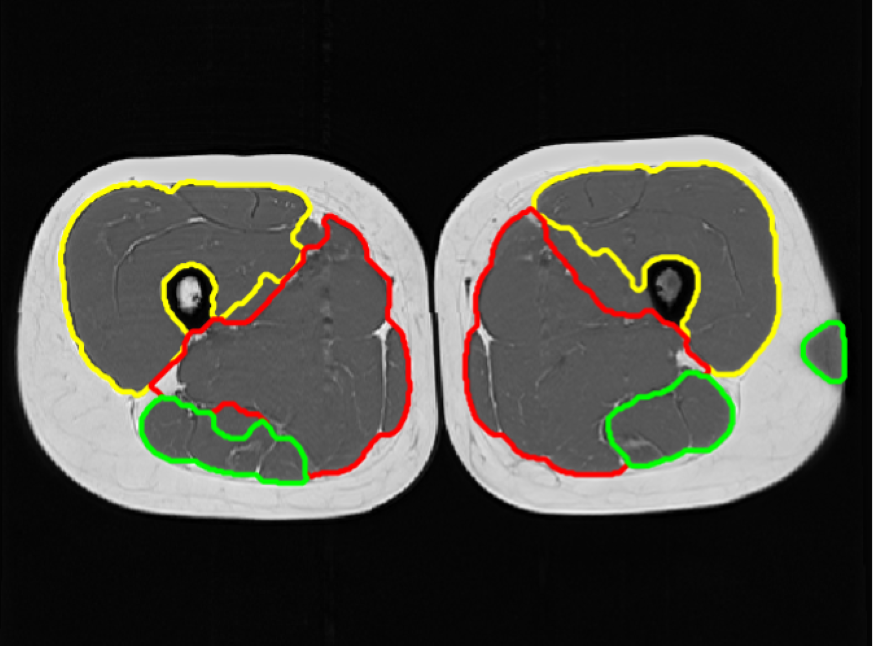}
   \includegraphics[height=5cm]{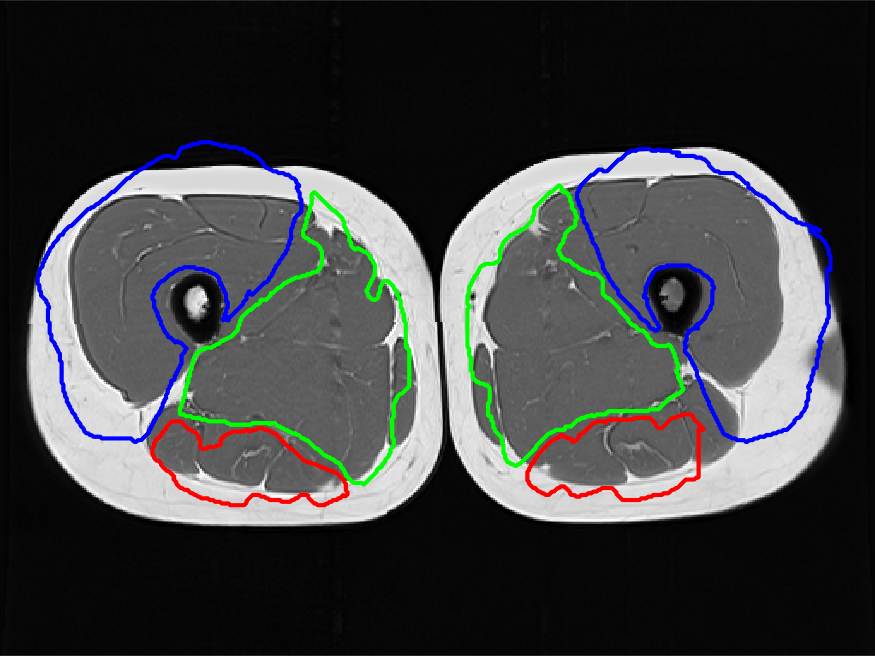}
   \includegraphics[height=5cm]{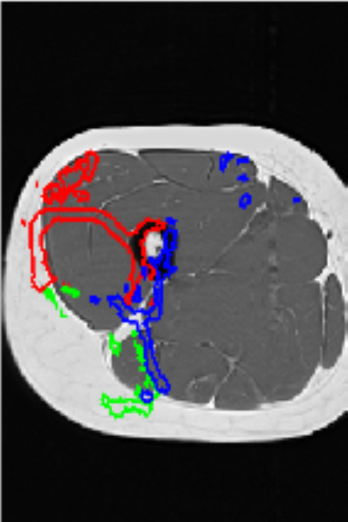}
   \end{tabular}
   \end{center}
   \caption[example] 
   { \label{fig:comp} 
Sample outputs from our proposed framework (left), M1 (center), and M2 (right)}
   \end{figure} 
 
   \begin{figure} [!htbp]
   \begin{center}
   \begin{tabular}{c} 
   \includegraphics[height=5cm]{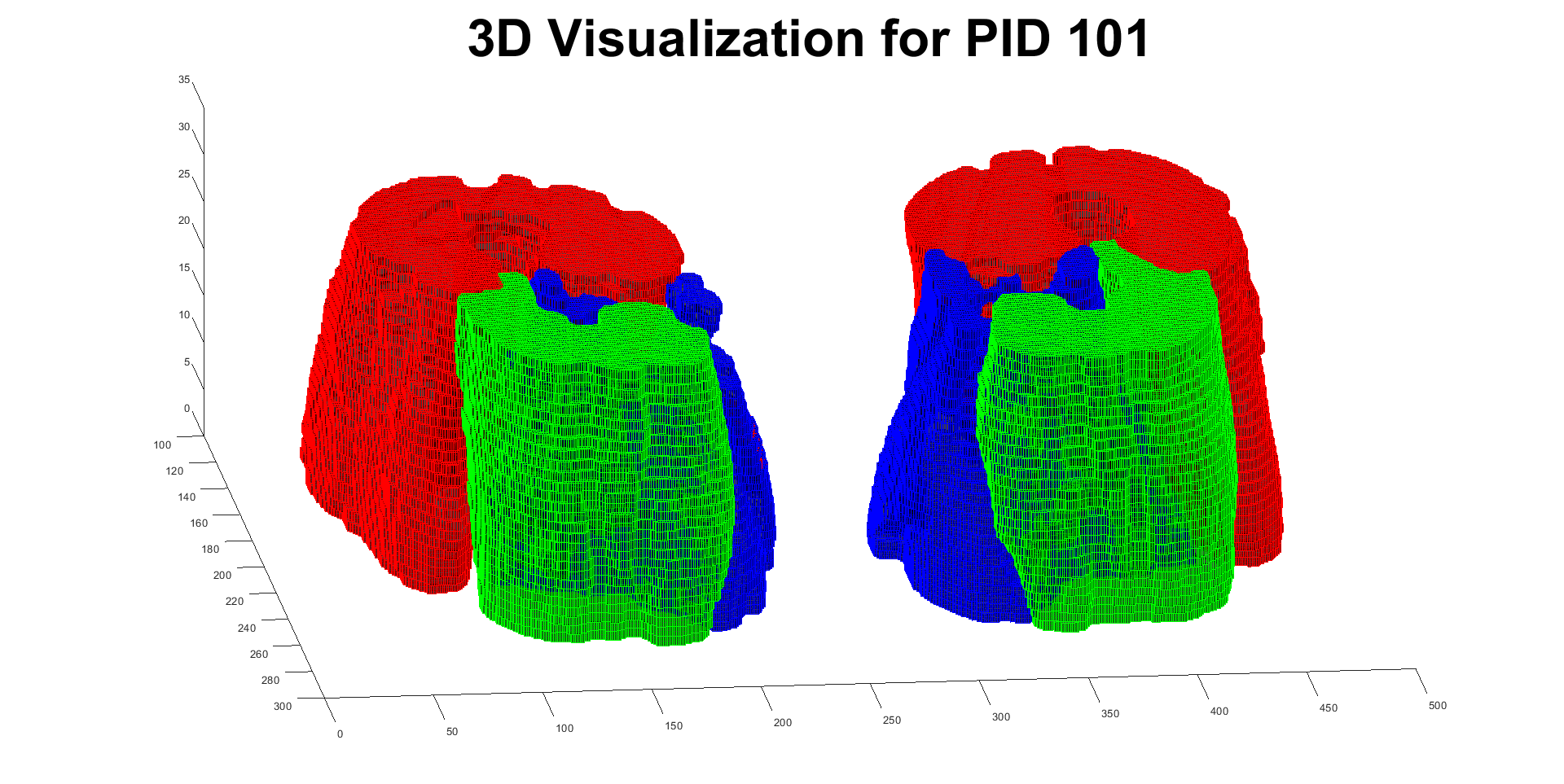}
   \end{tabular}
   \end{center}
   \caption[example] 
   { \label{fig:3d_vis} 
3D visualization of the a sample segmentation}
   \end{figure} 
We compare the results of our segmentation with two other related methods, namely M1 -- a non-parametric diffeomorphic image registration appraoch based on Thirion’s demons algorithm \cite{vercauteren2009diffeomorphic} and M2 -- a Joint Registration-Segmentation algorithm \cite{segreg2022}. The results for M1 and M2 have been outlined in Table \ref{tab:M1} and \ref{tab:M2} respectively. We also show some cross section comparison in Figure \ref{fig:comp} for one slice of a subject. Note that in M2 \cite{segreg2022}, two things are processed separately and we only show one of them. We can notice that our segmentation method outperforms M1 by a significant measure across all 3 muscle groups. Figure \ref{fig:3d_vis} shows a 3-D visualization for a sample segmentation from the proposed approach. It should be noted that our method is semi-supervised requiring no moving image, whereas M1 is completely automatic requiring one moving image that is preferably from the same dataset. We used PID 102 as the moving image while compiling the results for M1.

Comparisons with M2 led to similar conclusions, with our method outperforming M2 across all muscle groups. It should be noted that M2 performs exceptionally well when the moving image is chosen to be close to the fixed image. In fact, this algorithms would outperform our algorithm in such cases. However, in reality, having a moving image from the dataset may not always be feasible. Thus, we used a moving image from outside the dataset to compile the results for this method. Moreover, the results from M2 reflect how the performance of the algorithm may vary greatly depending on the nature of intensity correction in the subjects. Specifically, the algorihm performed sub optimally with patients 103-106 where the intensity correction was not perfect.

\section{CONCLUSION AND FUTURE WORK}

We proposed a semi-automatic local region of interest segmentation algorithm to handle images with  missing boundaries and intensity inhomogeneity. Our segmentation method starts with an initial contour (labeled by markers) inside the region of interest and evolves the contour using a geometric flow that incorporates  anti-markers as guidance for the stopping criterion. We also presented a new approach to remove intensity inhomogeneity, called PCBCFCM, that leverages a fat fraction image and fuzzy clustering. Numerical comparison with related work shows significantly better results. For future work, we also plan on using registration for placing the markers and anti-markers automatically.

\bibliographystyle{spiebib} 
\bibliography{Bib,references,Ref2} 

\begin{thebibliography}{10}

\bibitem{andrews2015generalized}
S.~Andrews and G.~Hamarneh, ``The generalized log-ratio transformation:
  learning shape and adjacency priors for simultaneous thigh muscle
  segmentation,'' {\em IEEE Transactions on Medical Imaging}~{\bf 34}(9),
  pp.~1773--1787, 2015.

\bibitem{baudin2012prior}
P.~Y. Baudin, N.~Azzabou, P.~G. Carlier, and N.~Paragios, ``Prior knowledge,
  random walks and human skeletal muscle segmentation,'' in {\em Medical Image
  Computing and Computer-Assisted Intervention},   {\bf 7510}, pp.~569--576,
  Springer, Berlin Heidelberg, 2012.

\bibitem{chen2003using}
Y.~Chen, W.~Guo, F.~Huang, D.~Wilson, and E.~A. Geiser, ``Using prior shape and
  points in medical image segmentation,'' in {\em Energy Minimization Methods
  in Computer Vision and Pattern Recognition},   {\bf 2683}, pp.~291--305,
  Springer, Berlin Heidelberg, 2003.

\bibitem{kemnitz2017validation}
J.~Kemnitz, F.~Eckstein, A.~G. Culvenor, A.~Ruhdorfer, T.~Dannhauer,
  S.~Ring-Dimitriou, A.~M. S{\"a}nger, and W.~Wirth, ``Validation of an active
  shape model-based semi-automated segmentation algorithm for the analysis of
  thigh muscle and adipose tissue cross-sectional areas,'' {\em Magnetic
  Resonance Materials in Physics, Biology and Medicine}~{\bf 30}(5),
  pp.~489--503, 2017.

\bibitem{lotjonen2010fast}
J.~M. L{\"o}tj{\"o}nen, R.~Wolz, J.~R. Koikkalainen, L.~Thurfjell, G.~Waldemar,
  H.~Soininen, D.~Rueckert, and A.~D.~N. Initiative, ``Fast and robust
  multi-atlas segmentation of brain magnetic resonance images,'' {\em
  Neuroimage}~{\bf 49}(3), pp.~2352--2365, 2010.

\bibitem{le2016volume}
A.~L. Troter, A.~Four{\'e}, M.~Guye, S.~Confort-Gouny, J.-P. Mattei, J.~Gondin,
  E.~Salort-Campana, and D.~Bendahan, ``Volume measurements of individual
  muscles in human quadriceps femoris using atlas-based segmentation
  approaches,'' {\em Magnetic Resonance Materials in Physics, Biology and
  Medicine}~{\bf 29}(2), pp.~245--257, 2016.

\bibitem{mesbah2019novel}
S.~Mesbah, A.~M. Shalaby, S.~Stills, A.~M. Soliman, A.~Willhite, S.~J. Harkema,
  E.~Rejc, and A.~S. El-Baz, ``Novel stochastic framework for automatic
  segmentation of human thigh {MRI} volumes and its applications in spinal cord
  injured individuals,'' {\em PloS One}~{\bf 14}(5), p.~e0216487, 2019.

\bibitem{sharma2019mammogram}
M.~K. Sharma, M.~Jas, V.~Karale, A.~Sadhu, and S.~Mukhopadhyay, ``Mammogram
  segmentation using multi-atlas deformable registration,'' {\em Computers in
  Biology and Medicine}~{\bf 110}, pp.~244--253, 2019.

\bibitem{yokota2018automated}
F.~Yokota, Y.~Otake, M.~Takao, T.~Ogawa, T.~Okada, N.~Sugano, and Y.~Sato,
  ``Automated muscle segmentation from {CT} images of the hip and thigh using a
  hierarchical multi-atlas method,'' {\em International Journal of Computer
  Assisted Radiology and Surgery}~{\bf 13}(7), pp.~977--986, 2018.

\bibitem{li2022image}
H.~Li, W.~Guo, J.~Liu, L.~Cui, and D.~Xie, ``Image segmentation with adaptive
  spatial priors from joint registration,'' {\em arXiv preprint
  arXiv:2203.15548} , 2022.

\bibitem{ahmad2014atlas}
E.~Ahmad, M.~H. Yap, H.~Degens, and J.~S. McPhee, ``Atlas-registration based
  image segmentation of {MRI} human thigh muscles in {3D} space,'' in {\em
  Medical Imaging 2014: Image Perception, Observer Performance, and Technology
  Assessment},   {\bf 9037}, pp.~424--435, SPIE, 2014.

\bibitem{jolivet2014skeletal}
E.~Jolivet, E.~Dion, P.~Rouch, G.~Dubois, R.~Charrier, C.~Payan, and W.~Skalli,
  ``Skeletal muscle segmentation from {MRI} dataset using a model-based
  approach,'' {\em Computer Methods in Biomechanics and Biomedical Engineering:
  Imaging and Visualization}~{\bf 2}(3), pp.~138--145, 2014.

\bibitem{molaie2020knowledge}
M.~Molaie and R.~A. Zoroofi, ``A knowledge-based modality-independent technique
  for concurrent thigh muscle segmentation: applicable to {CT} and {MR}
  images,'' {\em Journal of Digital Imaging}~{\bf 33}(5), pp.~1122--1135, 2020.

\bibitem{ogier2020novel}
A.~Ogier, L.~Heskamp, C.~P. Michel, A.~Four{\'e}, M.-E. Bellemare,
  A.~Le~Troter, A.~Heerschap, and D.~Bendahan, ``A novel segmentation framework
  dedicated to the follow-up of fat infiltration in individual muscles of
  patients with neuromuscular disorders,'' {\em Magnetic Resonance in
  Medicine}~{\bf 83}(5), pp.~1825--1836, 2020.

\bibitem{guomarker}
W.~Guo, M.~Judkovich, R.~Lartey, D.~Xie, M.~Yang, and X.~Li, ``A marker
  controlled active contour model for thigh muscle segmentation in mr images,''

\bibitem{ogier2017individual}
A.~Ogier, M.~Sdika, A.~Foure, A.~Le~Troter, and D.~Bendahan, ``Individual
  muscle segmentation in {MR} images: a {3D} propagation through {2D}
  non-linear registration approaches,'' in {\em 39th Annual International
  Conference of the IEEE Engineering in Medicine and Biology Society},
  pp.~317--320, IEEE, 2017.

\bibitem{chan2001active}
T.~F. Chan and L.~A. Vese, ``Active contours without edges,'' {\em IEEE
  Transactions on image processing}~{\bf 10}(2), pp.~266--277, 2001.

\bibitem{mumford1989optimal}
D.~Mumford and J.~Shah, ``Optimal approximations by piecewise smooth functions
  and associated variational problems,'' {\em Communications on pure and
  applied mathematics}~{\bf 42}(5), pp.~577--685, 1989.

\bibitem{kass1988snakes}
M.~Kass, A.~Witkin, and D.~Terzopoulos, ``Snakes: Active contour models,'' {\em
  International journal of computer vision}~{\bf 1}(4), pp.~321--331, 1988.

\bibitem{caselles1997geodesic}
V.~Caselles, R.~Kimmel, and G.~Sapiro, ``Geodesic active contours,'' {\em
  International journal of computer vision}~{\bf 22}(1), pp.~61--79, 1997.

\bibitem{gout2005segmentation}
C.~Gout, C.~Le~Guyader, and L.~Vese, ``Segmentation under geometrical
  conditions using geodesic active contours and interpolation using level set
  methods,'' {\em Numerical algorithms}~{\bf 39}(1-3), pp.~155--173, 2005.

\bibitem{spencer2015convex}
J.~Spencer and K.~Chen, ``A convex and selective variational model for image
  segmentation,'' {\em Communications in Mathematical Sciences}~{\bf 13}(6),
  pp.~1453--1472, 2015.

\bibitem{roberts2019convex}
M.~Roberts, K.~Chen, and K.~L. Irion, ``A convex geodesic selective model for
  image segmentation,'' {\em Journal of Mathematical Imaging and Vision}~{\bf
  61}(4), pp.~482--503, 2019.

\bibitem{wicks1993correction}
D.~A. Wicks, G.~J. Barker, and P.~S. Tofts, ``Correction of intensity
  nonuniformity in mr images of any orientation,'' {\em Magnetic resonance
  imaging}~{\bf 11}(2), pp.~183--196, 1993.

\bibitem{pham1999adaptive}
D.~L. Pham and J.~L. Prince, ``An adaptive fuzzy c-means algorithm for image
  segmentation in the presence of intensity inhomogeneities,'' {\em Pattern
  recognition letters}~{\bf 20}(1), pp.~57--68, 1999.

\bibitem{ahmed2002modified}
M.~N. Ahmed, S.~M. Yamany, N.~Mohamed, A.~A. Farag, and T.~Moriarty, ``A
  modified fuzzy c-means algorithm for bias field estimation and segmentation
  of mri data,'' {\em IEEE transactions on medical imaging}~{\bf 21}(3),
  pp.~193--199, 2002.

\bibitem{zosso2017image}
D.~Zosso, J.~An, J.~Stevick, N.~Takaki, M.~Weiss, L.~S. Slaughter, H.~H. Cao,
  P.~S. Weiss, and A.~L. Bertozzi, ``Image segmentation with dynamic artifacts
  detection and bias correction,'' {\em Inverse Problems and Imaging}~{\bf
  11}(3), pp.~577--600, 2017.

\bibitem{wells1996adaptive}
W.~M. Wells, W.~E.~L. Grimson, R.~Kikinis, and F.~A. Jolesz, ``Adaptive
  segmentation of mri data,'' {\em IEEE transactions on medical imaging}~{\bf
  15}(4), pp.~429--442, 1996.

\bibitem{liu2013image}
J.~Liu and H.~Zhang, ``Image segmentation using a local gmm in a variational
  framework,'' {\em Journal of mathematical imaging and vision}~{\bf 46}(2),
  pp.~161--176, 2013.

\bibitem{vercauteren2009diffeomorphic}
T.~Vercauteren, X.~Pennec, A.~Perchant, and N.~Ayache, ``Diffeomorphic demons:
  efficient non-parametric image registration,'' {\em NeuroImage}~{\bf 45}(1),
  pp.~S61--S72, 2009.

\bibitem{segreg2022}
H.~Li, W.~Guo, J.~Liu, L.~Cui, and D.~Xie, ``Image segmentation with adaptive
  spatial priors from joint registration,'' {\em SIAM J. IMAGING SCIENCES}~{\bf
  15}(3), pp.~1314--1344, 2022.

\bibitem{ghosh2017structured}
S.~Ghosh, N.~Ray, and P.~Boulanger, ``A structured deep-learning based approach
  for the automated segmentation of human leg muscle from {3D} {MRI},'' in {\em
  14th Conference on Computer and Robot Vision},  pp.~117--123, IEEE, 2017.

\bibitem{2020Clinical}
J.~Kemnitz, C.~F. Baumgartner, F.~Eckstein, A.~Chaudhari, A.~Ruhdorfer,
  W.~Wirth, S.~K. Eder, and E.~Konukoglu, ``Clinical evaluation of fully
  automated thigh muscle and adipose tissue segmentation using a {U-Net} deep
  learning architecture in context of osteoarthritic knee pain,'' {\em Magnetic
  Resonance Materials in Physics, Biology and Medicine}~{\bf 33}, pp.~483--493,
  2020.

\bibitem{ni2019automatic}
R.~Ni, C.~H. Meyer, S.~S. Blemker, J.~M. Hart, and X.~Feng, ``Automatic
  segmentation of all lower limb muscles from high-resolution magnetic
  resonance imaging using a cascaded three-dimensional deep convolutional
  neural network,'' {\em Journal of Medical Imaging}~{\bf 6}(4), p.~044009,
  2019.

\bibitem{guo2021fully}
Z.~Guo, H.~Zhang, Z.~Chen, E.~van~der Plas, L.~Gutmann, D.~Thedens,
  P.~Nopoulos, and M.~Sonka, ``Fully automated {3D} segmentation of {MR}-imaged
  calf muscle compartments: neighborhood relationship enhanced fully
  convolutional network,'' {\em Computerized Medical Imaging and Graphics}~{\bf
  87}, p.~101835, 2021.

\bibitem{burrows2020reproducible}
L.~Burrows, W.~Guo, K.~Chen, and F.~Torella, ``Reproducible kernel hilbert
  space based global and local image segmentation,'' {\em Inverse Problems \&
  Imaging}~{\bf 15}(1), p.~1, 2020.

\bibitem{deng2016single}
L.-J. Deng, W.~Guo, and T.-Z. Huang, ``Single image super-resolution by
  approximated heaviside functions,'' {\em Information Sciences}~{\bf 348},
  pp.~107--123, 2016.

\bibitem{zhao2005fast}
H.~Zhao, ``A fast sweeping method for eikonal equations,'' {\em Mathematics of
  computation}~{\bf 74}(250), pp.~603--627, 2005.

\bibitem{osher1988fronts}
S.~Osher and J.~A. Sethian, ``Fronts propagating with curvature-dependent
  speed: algorithms based on hamilton-jacobi formulations,'' {\em Journal of
  computational physics}~{\bf 79}(1), pp.~12--49, 1988.

\bibitem{guo2008geometric}
W.~Guo, Y.~Chen, and Q.~Zeng, ``A geometric flow-based approach for diffusion
  tensor image segmentation,'' {\em Philosophical Transactions of the Royal
  Society A: Mathematical, Physical and Engineering Sciences}~{\bf 366}(1874),
  pp.~2279--2292, 2008.

\end{thebibliography}

\end{document}